\def\papernamedversion{1}
\def\papersupplement{1}
\newcommand{\supplementsection}[1]{\section{#1}}
\RequirePackage[T1]{fontenc}
\documentclass[letterpaper,10pt,conference]{IEEEtran}
\usepackage{times}
\usepackage[numbers,sort&compress]{natbib}
\usepackage{graphicx}
\usepackage{capt-of}
\usepackage{caption} 
\usepackage{booktabs,array,tabularx}
\usepackage{amsmath,amssymb,bm}
\usepackage{xcolor,microtype}
\usepackage{algorithm}
\usepackage{algorithmic}
\usepackage{placeins}
\usepackage[hidelinks,breaklinks=true]{hyperref}
\IEEEoverridecommandlockouts
\AtBeginDocument{\bstctlcite{paperBibliographyStyle}}
\AtBeginDocument{\raggedbottom}
\newif\ifpaperanonymous
\ifdefined\papernamedversion
  \paperanonymousfalse
\else
  \paperanonymoustrue
\fi
\newcommand{\paperprojectleadnote}{Project leads.}
\newcommand{\papercorrespondingnote}{Corresponding author.}
\newcommand{\paperaffiliationone}{TAMS (Technical Aspects of Multimodal Systems), Department of Informatics, University of Hamburg, Hamburg, Germany.}
\newcommand{\paperaffiliationtwo}{The University of Hong Kong, Hong Kong, China.}
\newcommand{\paperaffiliationthree}{Technical University of Munich, Germany.}
\newcommand{\paperaffiliationfour}{Kinetix AI, China.}
\newcommand{\paperaffiliationfive}{University of Science and Technology of China, Hefei, China.}

\newcommand{\papernamedauthors}{%
  Ju Dong$^{1,\dagger}$,
  Xinyu Xiao$^{4}$,
  Yiming Zhou$^{4,\dagger}$,
  Tianshuo Yang$^{2}$,
  Xuerong Liu$^{2}$,\\
  Lei Zhang$^{1}$,
  Jian Chen$^{5}$,
  Haocheng Zhao$^{3}$,
  Yu Fu$^{1}$,
  Yujie Wei$^{4}$,
  Diwen Zheng$^{3}$,\\
  Alois Christian Knoll$^{3}$,
  Angela P. Schoellig$^{3}$,
  Ping Luo$^{2,*}$,
  Jianwei Zhang$^{1}$%
  \thanks{$^{\dagger}$\paperprojectleadnote\ $^{*}$\papercorrespondingnote}%
  \thanks{$^{1}$\paperaffiliationone}%
  \thanks{$^{2}$\paperaffiliationtwo}%
  \thanks{$^{3}$\paperaffiliationthree}%
  \thanks{$^{4}$\paperaffiliationfour}%
  \thanks{$^{5}$\paperaffiliationfive}%
}

\newcommand{\paperauthors}{%
  \ifpaperanonymous Anonymous Authors\else\papernamedauthors\fi
}

\newcommand{\papertitleen}{Coordinating Humanoid Loco-Manipulation\\through Action-Conditioned Latent Foresight}

\ifpaperanonymous
  \newcommand{\paperprojecturl}{\url{https://anonymous.4open.science/w/Action-Conditioned-Latent-Foresight/}}
\else
  \newcommand{\paperprojecturl}{\url{https://ju6276.github.io/Action-Grounded-Foresight/}}
\fi
\newif\ifdraftnotes
\draftnotesfalse
\newif\ifdraftcolors
\draftcolorsfalse
\newif\ifpapergallery
\papergalleryfalse
\definecolor{draftcolor}{RGB}{148,76,29}

\newcommand{\oursname}{Ours}
\newcommand{\featureshiftname}{Feature-shift variant}
\newcommand{\alignedname}{\oursname}
\newif\ifrealrobotlayoutplaceholder
\realrobotlayoutplaceholderfalse

\newcommand{\R}{\mathbb{R}}
\newcommand{\E}{\mathbb{E}}
\newcommand{\sg}{\operatorname{sg}}
\newcommand{\pool}{\operatorname{pool}}
\newcommand{\mse}{\operatorname{MSE}}
\newcommand{\dist}{\operatorname{dist}}
\newcommand{\norm}[1]{\left\lVert#1\right\rVert}
\DeclareMathOperator*{\argmax}{arg\,max}

\newcommand{\AlignedRealStandardMacro}{\pendingnum}
\newcommand{\AlignedRealGeneralMacro}{\pendingnum}
\newcommand{\AlignedMemory}{\pendingnum}
\newcommand{\AlignedHz}{\pendingnum}

\newcommand{\AlignedTrainable}{\pendingnum}

\newcommand{\EndpointEfficiencyRows}{%
\textbf{\alignedname} & \AlignedMemory & \AlignedTrainable & \AlignedHz \\
}

\newcommand{\ClosedLoopNoReconCells}{\pendingnum & \pendingnum & \pendingnum & \pendingnum & \pendingnum}
\newcommand{\ClosedLoopVisualCells}{\pendingnum & \pendingnum & \pendingnum & \pendingnum & \pendingnum}
\newcommand{\ClosedLoopDiffOneCells}{\pendingnum & \pendingnum & \pendingnum & \pendingnum & \pendingnum}
\newcommand{\ClosedLoopDiffTwoCells}{\pendingnum & \pendingnum & \pendingnum & \pendingnum & \pendingnum}
\newcommand{\ClosedLoopFactorCells}{\pendingnum & \pendingnum & \pendingnum & \pendingnum & \pendingnum}
\newcommand{\ClosedLoopJointCells}{\EndpointAdditionalAblationCells}

\newcommand{\SimpleBenchmarkRows}{%
$\Psi_0$~\cite{psi0} & \textbf{10}\,\textbar{}\,\textbf{10}\,\textbar{}\,6 & \textbf{10}\,\textbar{}\,\textbf{10}\,\textbar{}\,\textbf{10} & 7\,\textbar{}\,7\,\textbar{}\,\textbf{10} & \underline{7}\,\textbar{}\,5\,\textbar{}\,6 & \textbf{10}\,\textbar{}\,\textbf{10}\,\textbar{}\,\underline{8} & \textbf{10}\,\textbar{}\,\underline{9}\,\textbar{}\,\underline{9} & 85.6 \\
GR00T N1.6~\cite{gr00tn1} & \textbf{10}\,\textbar{}\,\textbf{10}\,\textbar{}\,7 & 7\,\textbar{}\,7\,\textbar{}\,6 & 1\,\textbar{}\,3\,\textbar{}\,3 & 0\,\textbar{}\,0\,\textbar{}\,0 & \underline{9}\,\textbar{}\,\underline{9}\,\textbar{}\,7 & 4\,\textbar{}\,4\,\textbar{}\,1 & 48.9 \\
$\pi_{0.5}$~\cite{pi05} & \underline{7}\,\textbar{}\,5\,\textbar{}\,1 & \textbf{10}\,\textbar{}\,\textbf{10}\,\textbar{}\,8 & 5\,\textbar{}\,4\,\textbar{}\,5 & 3\,\textbar{}\,3\,\textbar{}\,3 & \textbf{10}\,\textbar{}\,\textbf{10}\,\textbar{}\,\underline{8} & 0\,\textbar{}\,0\,\textbar{}\,0 & 51.1 \\
InternVLA-M1~\cite{internvlam1} & 0\,\textbar{}\,0\,\textbar{}\,0 & 5\,\textbar{}\,5\,\textbar{}\,0 & 0\,\textbar{}\,0\,\textbar{}\,0 & 0\,\textbar{}\,0\,\textbar{}\,0 & 0\,\textbar{}\,0\,\textbar{}\,0 & 3\,\textbar{}\,5\,\textbar{}\,7 & 13.9 \\
H-RDT~\cite{hrdt} & 0\,\textbar{}\,0\,\textbar{}\,2 & 0\,\textbar{}\,0\,\textbar{}\,1 & 0\,\textbar{}\,1\,\textbar{}\,0 & 0\,\textbar{}\,0\,\textbar{}\,0 & 0\,\textbar{}\,0\,\textbar{}\,0 & 0\,\textbar{}\,0\,\textbar{}\,0 & 2.2 \\
DreamZero~\cite{dreamzero} & \textbf{10}\,\textbar{}\,\textbf{10}\,\textbar{}\,\textbf{10} & \underline{9}\,\textbar{}\,\underline{9}\,\textbar{}\,8 & 7\,\textbar{}\,8\,\textbar{}\,\underline{9} & 5\,\textbar{}\,3\,\textbar{}\,3 & \underline{9}\,\textbar{}\,\textbf{10}\,\textbar{}\,7 & 0\,\textbar{}\,0\,\textbar{}\,1 & 65.6 \\
EgoVLA~\cite{egovla} & 0\,\textbar{}\,1\,\textbar{}\,2 & 7\,\textbar{}\,5\,\textbar{}\,8 & 0\,\textbar{}\,4\,\textbar{}\,3 & 0\,\textbar{}\,0\,\textbar{}\,0 & \textbf{10}\,\textbar{}\,\textbf{10}\,\textbar{}\,7 & 3\,\textbar{}\,5\,\textbar{}\,4 & 38.3 \\
DP~\cite{diffusionpolicy} & 3\,\textbar{}\,3\,\textbar{}\,2 & \textbf{10}\,\textbar{}\,8\,\textbar{}\,6 & 3\,\textbar{}\,2\,\textbar{}\,4 & 4\,\textbar{}\,0\,\textbar{}\,0 & 8\,\textbar{}\,\underline{9}\,\textbar{}\,\underline{8} & 0\,\textbar{}\,0\,\textbar{}\,0 & 38.9 \\
ACT~\cite{act} & \textbf{10}\,\textbar{}\,\textbf{10}\,\textbar{}\,5 & \textbf{10}\,\textbar{}\,\underline{9}\,\textbar{}\,\underline{9} & 7\,\textbar{}\,7\,\textbar{}\,\textbf{10} & 5\,\textbar{}\,5\,\textbar{}\,5 & \textbf{10}\,\textbar{}\,\textbf{10}\,\textbar{}\,\underline{8} & \underline{9}\,\textbar{}\,\textbf{10}\,\textbar{}\,\textbf{10} & 82.8 \\
DiT4DiT~\cite{dit4dit} & \textbf{10}\,\textbar{}\,\underline{9}\,\textbar{}\,\underline{9} & \underline{9}\,\textbar{}\,\underline{9}\,\textbar{}\,\textbf{10} & \textbf{9}\,\textbar{}\,\underline{9}\,\textbar{}\,\textbf{10} & \underline{7}\,\textbar{}\,\underline{7}\,\textbar{}\,\underline{7} & \textbf{10}\,\textbar{}\,\underline{9}\,\textbar{}\,7 & \underline{9}\,\textbar{}\,\underline{9}\,\textbar{}\,\underline{9} & \underline{87.8} \\
Fast-WAM~\cite{fastwam} & \textbf{10}\,\textbar{}\,\textbf{10}\,\textbar{}\,6 & 0\,\textbar{}\,3\,\textbar{}\,1 & 6\,\textbar{}\,5\,\textbar{}\,7 & 0\,\textbar{}\,0\,\textbar{}\,0 & \textbf{10}\,\textbar{}\,7\,\textbar{}\,7 & 0\,\textbar{}\,0\,\textbar{}\,0 & 40.0 \\
VLA-JEPA~\cite{vlajepa} & 0\,\textbar{}\,0\,\textbar{}\,0 & 1\,\textbar{}\,0\,\textbar{}\,0 & 7\,\textbar{}\,8\,\textbar{}\,\underline{9} & \underline{7}\,\textbar{}\,\underline{7}\,\textbar{}\,\textbf{9} & \underline{9}\,\textbar{}\,8\,\textbar{}\,5 & 0\,\textbar{}\,3\,\textbar{}\,4 & 42.8 \\%
}

\newcommand{\SimpleAdditionalRows}{%
$\Psi_0$~\cite{psi0} & 7\,\textbar{}\,5\,\textbar{}\,4 & 7\,\textbar{}\,6\,\textbar{}\,3 & 3\,\textbar{}\,3\,\textbar{}\,4 & \underline{7}\,\textbar{}\,\textbf{9}\,\textbar{}\,\textbf{9} & 55.8 & 73.7 \\
DiT4DiT~\cite{dit4dit} & \textbf{10}\,\textbar{}\,\textbf{10}\,\textbar{}\,\underline{9} & 7\,\textbar{}\,\underline{8}\,\textbar{}\,\textbf{8} & \underline{5}\,\textbar{}\,\underline{6}\,\textbar{}\,\textbf{7} & 6\,\textbar{}\,7\,\textbar{}\,\underline{5} & \underline{73.3} & \underline{82.0} \\
Fast-WAM~\cite{fastwam} & 1\,\textbar{}\,1\,\textbar{}\,0 & 0\,\textbar{}\,0\,\textbar{}\,0 & 2\,\textbar{}\,1\,\textbar{}\,0 & 2\,\textbar{}\,2\,\textbar{}\,1 & 8.3 & 27.3 \\
VLA-JEPA~\cite{vlajepa} & 8\,\textbar{}\,\underline{9}\,\textbar{}\,7 & \underline{8}\,\textbar{}\,\underline{8}\,\textbar{}\,\underline{7} & \textbf{6}\,\textbar{}\,2\,\textbar{}\,4 & 4\,\textbar{}\,3\,\textbar{}\,4 & 58.3 & 49.0 \\%
}

\newcommand{\EndpointSimpleRows}{%
\textbf{\alignedname} & \textbf{10}\,\textbar{}\,\textbf{10}\,\textbar{}\,7 & \textbf{10}\,\textbar{}\,\underline{9}\,\textbar{}\,\textbf{10} & \underline{8}\,\textbar{}\,\textbf{10}\,\textbar{}\,\underline{9} & \textbf{10}\,\textbar{}\,\textbf{10}\,\textbar{}\,\textbf{9} & \textbf{10}\,\textbar{}\,\textbf{10}\,\textbar{}\,\textbf{10} & \textbf{10}\,\textbar{}\,\textbf{10}\,\textbar{}\,\textbf{10} & \textbf{95.6} \\%
}

\newcommand{\EndpointAdditionalRows}{%
\textbf{\alignedname} & \underline{9}\,\textbar{}\,\textbf{10}\,\textbar{}\,\textbf{10} & \textbf{9}\,\textbar{}\,\textbf{9}\,\textbar{}\,\textbf{8} & \underline{5}\,\textbar{}\,\textbf{7}\,\textbar{}\,\underline{5} & \textbf{8}\,\textbar{}\,\underline{8}\,\textbar{}\,\textbf{9} & \textbf{80.8} & \textbf{89.7} \\%
}

\newcommand{\EndpointAdditionalAblationCells}{%
9\,\textbar{}\,10\,\textbar{}\,10 & 9\,\textbar{}\,9\,\textbar{}\,8 & 5\,\textbar{}\,7\,\textbar{}\,5 & 8\,\textbar{}\,8\,\textbar{}\,9 & 80.8%
}

\newcommand{\EndpointFaucetSuccesses}{%
17%
}

\newcommand{\EndpointFaucetTrials}{%
30%
}

\newcommand{\EndpointSimulationFailures}{%
31%
}

\newcommand{\EndpointFaucetFailures}{%
13%
}

\newcommand{\EndpointFaucetFailureShare}{%
41.9%
}

\newcommand{\SimplePredictionRows}{%
\featureshiftname & 91.7 & 73.3 & 84.3 \\
\textbf{\alignedname} & \textbf{95.6} & \textbf{80.8} & \textbf{89.7} \\%
}

\newcommand{\SimpleEfficiencyBaselineRows}{%
GR00T N1.6~\cite{gr00tn1} & 6.8 & 1.62 & 15.46 \\
DiT4DiT~\cite{dit4dit} & 21 & 2.35 & 3.8 \\
Fast-WAM~\cite{fastwam} & 28 & 6.7 & 5.2 \\%
}

\newcommand{\OrinMemoryMean}{12.84}

\newcommand{\OrinHz}{0.73}

\newcommand{\AlignedSimMacro}{89.7}

\newcommand{\AlignedSimMainMacro}{95.6}

\newcommand{\AlignedSimAdditionalMacro}{80.8}

\newcommand{\SimpleOursMacro}{84.3}

\newcommand{\SimpleVLAMacro}{49.0}

\newcommand{\SimpleDitMacro}{82.0}

\newcommand{\SimplePsiMacro}{73.7}

\newcommand{\TrainingGPUCount}{8}
\newcommand{\SimTrainingSteps}{40,000}
\newcommand{\RealTrainingSteps}{90,000}
\newcommand{\SimBatchPerGPU}{32}
\newcommand{\RealBatchPerGPU}{4}

\newcommand{\DataEfficiencyPSIHundred}{56.7}
\newcommand{\DataEfficiencyOursHundred}{84.0}
\newcommand{\DataEfficiencyHundredGain}{27.3}

\providecommand{\AlignedMemory}{}
\renewcommand{\AlignedMemory}{7.2}
\providecommand{\AlignedHz}{}
\renewcommand{\AlignedHz}{9.1}
\providecommand{\AlignedTrainable}{}
\renewcommand{\AlignedTrainable}{3.0}

\providecommand{\ConfirmedRealStandardRows}{}
\renewcommand{\ConfirmedRealStandardRows}{%
GR00T N1.7 & \textbf{9/10} & \underline{6/10} & \underline{8/10} & \textbf{8/10} & 2/10 & \underline{8/10} & \underline{8/10} & \underline{70.0} \\
$\Psi_0$ & 2/10 & \underline{6/10} & 6/10 & 4/10 & 2/10 & 2/10 & 1/10 & 32.9 \\
DiT4DiT & 4/10 & \textbf{8/10} & \underline{8/10} & \underline{6/10} & \underline{8/10} & 4/10 & 6/10 & 62.9 \\
\textbf{\oursname} & \underline{6/10} & \underline{6/10} & \textbf{10/10} & \textbf{8/10} & \textbf{10/10} & \textbf{10/10} & \textbf{10/10} & \textbf{85.7} \\
}
\providecommand{\ConfirmedRealGeneralizationRows}{}
\renewcommand{\ConfirmedRealGeneralizationRows}{%
GR00T N1.7 & \textbf{4/10} & \textbf{4/10} & \underline{5/10} & \underline{5/10} & \underline{6/10} & \underline{4/10} & \underline{6/10} & \underline{48.6} \\
$\Psi_0$ & \underline{2/10} & \underline{3/10} & 0/10 & 3/10 & 3/10 & 2/10 & 2/10 & 21.4 \\
DiT4DiT & 1/10 & \textbf{4/10} & \underline{5/10} & \underline{5/10} & 4/10 & 3/10 & 5/10 & 38.6 \\
\textbf{\oursname} & \textbf{4/10} & \textbf{4/10} & \textbf{8/10} & \textbf{6/10} & \textbf{7/10} & \textbf{8/10} & \textbf{9/10} & \textbf{65.7} \\
}
\providecommand{\AlignedRealStandardMacro}{}
\renewcommand{\AlignedRealStandardMacro}{85.7}
\providecommand{\AlignedRealGeneralMacro}{}
\renewcommand{\AlignedRealGeneralMacro}{65.7}
\providecommand{\LayoutRealGRStandardMacro}{}
\renewcommand{\LayoutRealGRStandardMacro}{70.0}
\providecommand{\LayoutRealGRGeneralMacro}{}
\renewcommand{\LayoutRealGRGeneralMacro}{48.6}

\providecommand{\LayoutRealStandardGain}{}
\renewcommand{\LayoutRealStandardGain}{15.7}

\providecommand{\ClosedLoopNoReconCells}{}
\renewcommand{\ClosedLoopNoReconCells}{7\,\textbar{}\,7\,\textbar{}\,6 & 7\,\textbar{}\,6\,\textbar{}\,6 & 3\,\textbar{}\,4\,\textbar{}\,3 & 6\,\textbar{}\,6\,\textbar{}\,5 & 55.0}
\providecommand{\ClosedLoopVisualCells}{}
\renewcommand{\ClosedLoopVisualCells}{8\,\textbar{}\,8\,\textbar{}\,8 & 8\,\textbar{}\,7\,\textbar{}\,7 & 4\,\textbar{}\,5\,\textbar{}\,4 & 7\,\textbar{}\,6\,\textbar{}\,7 & 65.8}
\providecommand{\ClosedLoopDiffOneCells}{}
\renewcommand{\ClosedLoopDiffOneCells}{8\,\textbar{}\,8\,\textbar{}\,8 & 8\,\textbar{}\,8\,\textbar{}\,7 & 4\,\textbar{}\,5\,\textbar{}\,4 & 7\,\textbar{}\,7\,\textbar{}\,7 & 67.5}
\providecommand{\ClosedLoopDiffTwoCells}{}
\renewcommand{\ClosedLoopDiffTwoCells}{8\,\textbar{}\,9\,\textbar{}\,8 & 8\,\textbar{}\,7\,\textbar{}\,8 & 4\,\textbar{}\,6\,\textbar{}\,5 & 7\,\textbar{}\,7\,\textbar{}\,7 & 70.0}
\providecommand{\ClosedLoopFactorCells}{}
\renewcommand{\ClosedLoopFactorCells}{8\,\textbar{}\,9\,\textbar{}\,9 & 8\,\textbar{}\,8\,\textbar{}\,8 & 4\,\textbar{}\,6\,\textbar{}\,4 & 8\,\textbar{}\,7\,\textbar{}\,7 & 71.7}
\providecommand{\AblationNoReconSuccesses}{}
\renewcommand{\AblationNoReconSuccesses}{66}
\providecommand{\AblationNoReconPercent}{}
\renewcommand{\AblationNoReconPercent}{55.0}
\providecommand{\AblationVisualSuccesses}{}
\renewcommand{\AblationVisualSuccesses}{79}
\providecommand{\AblationVisualPercent}{}
\renewcommand{\AblationVisualPercent}{65.8}

\providecommand{\AblationDiffOnePercent}{}
\renewcommand{\AblationDiffOnePercent}{67.5}

\providecommand{\AblationDiffTwoPercent}{}
\renewcommand{\AblationDiffTwoPercent}{70.0}

\providecommand{\AblationJointSuccesses}{}
\renewcommand{\AblationJointSuccesses}{97}
\providecommand{\AblationJointPercent}{}
\renewcommand{\AblationJointPercent}{80.8}
\providecommand{\AblationReconGain}{}
\renewcommand{\AblationReconGain}{10.8}
\providecommand{\AblationJointVisualGain}{}
\renewcommand{\AblationJointVisualGain}{15.0}
\providecommand{\AblationJointFactorGain}{}
\renewcommand{\AblationJointFactorGain}{9.2}

\usepackage{float,flafter,stfloats}

\title{\LARGE\bfseries \papertitleen}
\author{\paperauthors}

\begin{document}
\makeatletter
\let\@oldmaketitle\@maketitle
\renewcommand{\@maketitle}{\@oldmaketitle
  \centering
  \includegraphics[width=\textwidth]{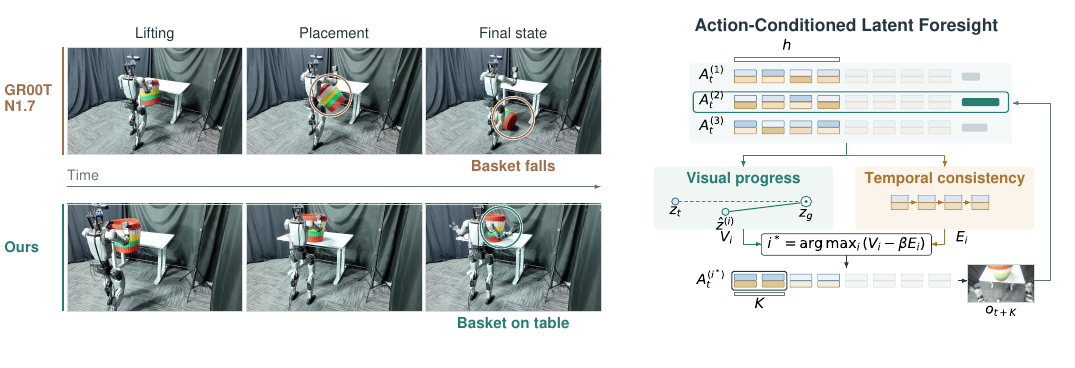}
  \captionof{figure}{Left: two independent Place Basket on Table recordings with GR00T N1.7 and our method. Right: each sampled sequence $A_t^{(i)}$ is scored on its first $h$ commands. Visual progress $V_i$ is the predicted reduction in distance to the subgoal $z_g$ from current features $z_t$ to predicted features $\hat z^{(i)}$; temporal-consistency energy $E_i$ measures next-command prediction errors. The robot executes the first $K$ commands of the best-scoring sequence $i^*$ and replans from $o_{t+K}$.}
  \label{fig:teaser}\vspace{5pt}}
\makeatother
\maketitle
\setcounter{figure}{1}
\thispagestyle{empty}\pagestyle{empty}

\begin{abstract}
Humanoid loco-manipulation requires coordinated body motion and hand interaction over time, and a plausible-looking command sequence can still drop a held object or leave an interaction incomplete. We present action-conditioned latent foresight, a latent world--action model (WAM) that evaluates proposed command sequences before execution. An action generator proposes joint motion--hand command sequences from vision, language, and the robot state. An action-conditioned endpoint predictor estimates the visual features each sequence prefix reaches and measures progress toward a demonstration-derived subgoal, while a joint temporal prior scores temporal consistency given the robot state and preceding commands. The robot executes the first commands of the best-scoring sequence and replans from the next observation. During training, action consistency regularization reconstructs the first demonstrated command from observed and predicted feature changes, encouraging the predictor to retain action-relevant information. On ten tasks in the humanoid loco-manipulation simulation benchmark SIMPLE, our method achieves \AlignedSimMacro\% average success, compared with \SimplePsiMacro\% for $\Psi_0$ and \SimpleDitMacro\% for DiT4DiT. On seven real-robot tasks, it achieves \AlignedRealStandardMacro\% under standard conditions and \AlignedRealGeneralMacro\% under unseen object configurations, while GR00T N1.7 achieves \LayoutRealGRStandardMacro\% and \LayoutRealGRGeneralMacro\%, respectively. Project page: \paperprojecturl.
\end{abstract}

\section{Introduction}\label{sec:introduction}
Everyday humanoid tasks require coordinated actions over time. To place a basket on a table, the robot must raise and position the basket while maintaining its grasp until the table supports it. To load a washer, it must open the door, carry a T-shirt to the drum without dropping it, and release it only once it is inside. These tasks connect whole-body motion with hand interaction: an execution decision concerns both the scene a command sequence will reach and the relationships among commands along the way.

Vision-language-action (VLA) policies generate commands from images and language instructions~\cite{openvla,pi0,pi05}, while humanoid models incorporate human motion and predictive representations~\cite{psi0,wholebodyvla,motionwam,omega0}. When an action generator proposes several sequences, the robot must decide which one to execute. Latent world models provide predictions for goal-directed planning~\cite{dinowm,vjepa2}. Most world--action models use future prediction to improve command generation~\cite{dreamzero,dit4dit,fastwam} rather than to compare sampled sequences before execution. Existing verifiers use VLM judgments~\cite{robomonkey,forewarn}, while DreamTrajectory combines predicted trajectory agreement with action smoothness~\cite{dreamtrajectory}. Two signals are relevant for evaluating proposed humanoid actions: predicted visual progress evaluates the endpoint relative to a task goal, while temporal consistency evaluates relationships among commands within the sequence.

We introduce \emph{action-conditioned latent foresight}, a latent world--action model (WAM) that uses both signals for humanoid loco-manipulation (Fig.~\ref{fig:teaser}). An action generator proposes joint motion--hand command sequences; an action-conditioned endpoint predictor scores their visual progress toward a demonstration-derived subgoal, and a joint temporal prior scores their temporal consistency. The robot executes the first commands of the best-scoring sequence and replans from the next observation.

Our contributions are threefold:
\begin{itemize}
  \item We propose action-conditioned latent foresight, which scores generated motion--hand sequences before execution using predicted visual progress and a learned joint temporal prior.
  \item We develop action consistency regularization for latent prediction. A training-only inverse action decoder reconstructs the first demonstrated command from both observed and predicted feature changes, encouraging the predicted features to retain information about the executed action.
  \item We evaluate ten SIMPLE~\cite{simple} simulation tasks and seven real-robot tasks, including unseen object configurations. Our method exceeds the evaluated baselines in average success; closed-loop ablations show gains from action consistency regularization and joint temporal scoring.
\end{itemize}

\section{Related Work}\label{sec:related}
\subsection{Humanoid VLAs and World--Action Models}
VLA policies transfer vision--language representations to command generation~\cite{openvla,pi0}. Heterogeneous co-training improves open-world generalization~\cite{pi05}, and diffusion action heads generate continuous action chunks~\cite{gr00tn1}. For humanoids, EgoVLA uses retargeted wrist and hand motion~\cite{egovla}, $\Psi_0$ exploits egocentric pretraining~\cite{psi0}, and WholeBodyVLA learns manipulation and locomotion latent actions from video before decoding upper-body and locomotion commands~\cite{wholebodyvla}.

World--action models couple prediction with command generation. DreamZero jointly models video and actions~\cite{dreamzero}; DiT4DiT and MotionWAM condition action or motion diffusion on intermediate video-model features~\cite{dit4dit,motionwam}. MotionWAM combines SONIC motion codes with hand-control channels~\cite{motionwam}, while $\omega$-0 generates controller-compatible actions from future embeddings~\cite{omega0}. WholeBodyWAM pretrains a motion expert to learn whole-body motion priors for robot action generation~\cite{wholebodymotion}. Fast-WAM and InternW0-$\Delta$ use future supervision during training without generating video at test time~\cite{fastwam,internw0}. In our method, the action generator supplies command sequences, and each sequence prefix is scored before execution using action-conditioned prediction and temporal consistency.

\subsection{Action Supervision for Latent Prediction}
Latent world models predict action-conditioned feature transitions for goal-directed planning~\cite{dinowm,vjepa2}. V-JEPA 2.1 learns dense video features~\cite{vjepa21}, and a study of JEPA world models examines how representation, predictor, and planner choices affect physical planning~\cite{jepawms}. VLA-JEPA learns VLM latent tokens through future-state prediction and uses them to condition flow-matching action generation~\cite{vlajepa}. Related models combine feature prediction with action modeling~\cite{jepawam}, or predict future latent states alongside actions~\cite{lewam}. Our predictor takes the proposed command prefix as input and estimates its endpoint features, allowing command sequences sampled from the same observation to receive different visual scores.

Action supervision can also arise from discrete or constrained continuous latents~\cite{lapa,lawm}, inverse dynamics that turn generated video into executable actions~\cite{unipi}, or visual correspondences~\cite{avdc}. Our action consistency regularization uses an inverse action decoder to reconstruct the first demonstrated command from observed and predicted feature changes, providing a training objective for the endpoint predictor rather than a test-time action decoder.

\subsection{Temporal Action Priors and Sequence Scoring}
Action-sequence models~\cite{diffusionpolicy,act} and cross-embodiment action priors~\cite{actionpriors} learn regularities that guide generation. PRISM combines state-conditioned priors with planning distributions~\cite{prism}, and DynaGuide uses learned dynamics to steer diffusion~\cite{dynaguide}. Action verification evaluates proposed actions before execution: RoboMonkey learns action preferences with a VLM verifier~\cite{robomonkey}, while FOREWARN aligns predicted latent rollouts with behavior descriptions for VLM-based action choice~\cite{forewarn}.

DreamTrajectory combines agreement with intended end-effector trajectories and fixed smoothness penalties~\cite{dreamtrajectory}; LeWAM's DemoDPO uses demonstration preferences for offline policy refinement~\cite{lewam}. Our temporal prior learns to predict the next complete command from the initial robot state and preceding motion--hand commands. Its prediction error is combined with visual progress when scoring proposed actions.

\section{Method}\label{sec:method}
\begin{figure*}[t]
\centering
\includegraphics[width=\textwidth]{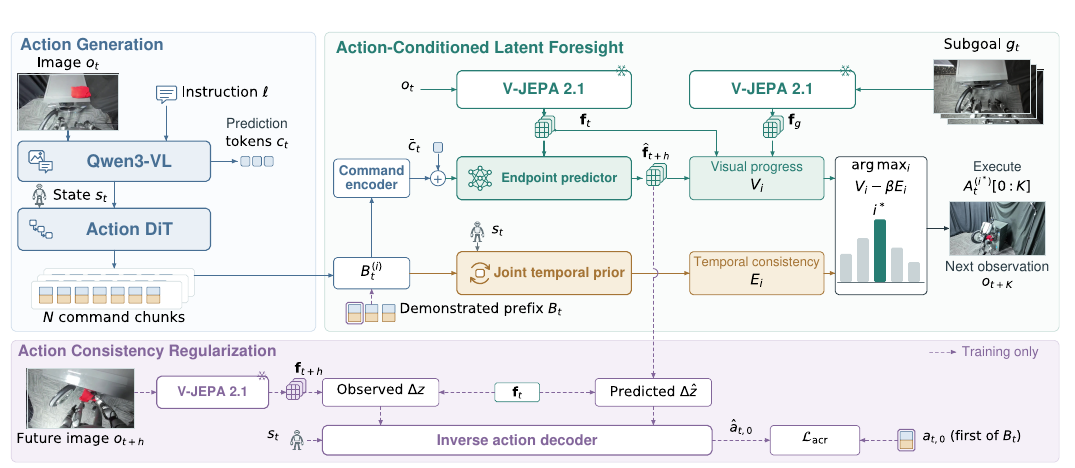}
\caption{Architecture. $\mathbf f$ denotes V-JEPA features. Each prefix $B_t^{(i)}$ receives visual progress $V_i$ toward the active subgoal $g_t$ (front of the stack) and temporal-consistency energy $E_i$. In training, the demonstrated prefix $B_t$ replaces sampled prefixes, and its first command $a_{t,0}$ is reconstructed as $\hat a_{t,0}$ from $\Delta z$ and from $\Delta\hat z$. Score bars are schematic; snowflakes: frozen weights.}
\label{fig:supervision}
\end{figure*}

\subsection{Problem Formulation and Receding-Horizon Scoring}
\label{sec:decision}
Given an egocentric image $o_t$, instruction $\ell$, and proprioceptive state $s_t$, our action generator samples $N$ normalized command chunks $A_t^{(i)}\in\R^{H\times d_a}$:
\begin{equation}
 A_t^{(i)}\sim\pi_\omega(\cdot\mid x_t),\quad
 A_t^{(i)}=(a_{t,0}^{(i)},\ldots,a_{t,H-1}^{(i)}),
 \label{eq:proposals}
\end{equation}
where $x_t=(o_t,\ell,s_t)$. Each command contains body-motion components $m_k$ and hand components $u_k$ defined by the controller interface, with $d_a=d_m+d_u$. On the real robot, $m_k$ is a SONIC motion latent~\cite{sonic} and $u_k$ contains hand joint targets.

Let $H$ be the proposal length, $h$ the evaluation horizon, and $K$ the executed prefix length, with $1\leq K\leq h\leq H$ and $h\geq2$. The endpoint predictor and the temporal prior both score the prefix $B_t^{(i)}=A_t^{(i)}[0:h]$. Progress is measured toward temporally ordered visual subgoals $G=(g_1,\ldots,g_J)$ extracted from successful demonstrations, with active subgoal $g_t$.

Let $z_t$ and $z_g$ be normalized current and goal features, $\widehat{\mathbf f}_{t+h}^{(i)}$ the predicted endpoint features, and $\pool$ spatial mean pooling followed by normalization, so that $\hat z^{(i)}=\pool(\widehat{\mathbf f}_{t+h}^{(i)})$. With cosine distance $\dist$, each sequence receives a score that combines predicted visual progress with temporal-consistency energy:
\begin{align}
 V_i&=\dist(z_t,z_g)-\dist(\hat z^{(i)},z_g),\nonumber\\
 i^*&=\argmax_i\{V_i-\beta E_i\},\quad E_i=E_\phi(B_t^{(i)};s_t).
 \label{eq:score}
\end{align}
In a receding-horizon loop, the robot executes the first $K$ commands of $A_t^{(i^*)}$ and repeats the decision from the next observation; $h>K$ provides lookahead beyond the next decision. A subgoal tracker advances $g_t$ to the next subgoal in $G$ when $\dist(z_t,z_g)<\epsilon_g$.

\subsection{Action Generation}
\label{sec:generator}
Qwen3-VL-2B~\cite{qwen3vl} encodes the image and instruction into separate policy tokens $c_t^{\mathrm{act}}$ and prediction tokens $c_t$ (Fig.~\ref{fig:supervision}). Action DiT~\cite{dit} alternates cross-attention to $c_t^{\mathrm{act}}$ with self-attention over state tokens, learned tokens, and noisy action tokens. Its flow-matching velocity head~\cite{flowmatching} generates command sequences through numerical integration from initial noise $\xi$.

\subsection{Action-Conditioned Endpoint Dynamics}
\label{sec:predictor}
We fine-tune the V-JEPA~2.1 ViT-L/16 encoder (384~px)~\cite{vjepa21} with its official self-supervised training recipe on HIW-500~\cite{hiw500_2026}, over 500 hours of whole-body humanoid demonstrations collected through Unitree G1 teleoperation in real homes. We then freeze the resulting encoder for task training and inference; it also extracts the demonstration subgoals. Let $\Phi$ denote this encoder applied to a single image, repeated to the encoder clip length, with its last temporal feature block retained. Current, goal, and future images are encoded separately:
\begin{equation}
 \mathbf f_t=\Phi(o_t),\quad \mathbf f_g=\Phi(g_t),\quad \mathbf f_{t+h}=\Phi(o_{t+h}).
 \label{eq:endpoints}
\end{equation}
Thus $z_t=\pool(\mathbf f_t)$ and $z_g=\pool(\mathbf f_g)$. During training, the demonstrated prefix $B_t=(a_{t,0},\ldots,a_{t,h-1})$ starting at time $t$ is paired with the observation $o_{t+h}$.

A command encoder $C_\eta$ produces one token per command with positional encoding. We add the mean of the prediction tokens, $\bar c_t$, to each command token, giving $q_t(B)=C_\eta(B)+\bar c_t$. The endpoint predictor is
\begin{equation}
 \widehat{\mathbf f}_{t+h}=\mathcal P_\theta(\mathbf f_t,q_t(B_t)).
 \label{eq:prediction}
\end{equation}

\begingroup\interlinepenalty=10000
\subsection{Joint Temporal Prior}
\label{sec:prior}
A shared GRU predicts the next complete motion--hand command from the initial robot state and preceding commands:
\begin{align}
 r_k&=\operatorname{GRU}_\phi(r_{k-1},e_\phi([s_t;m_{k-1};u_{k-1}])),\nonumber\\
 [\tilde m_k;\tilde u_k]&=\operatorname{MLP}_\phi(r_k),\qquad k=1,\ldots,h-1.
 \label{eq:prior}
\end{align}
\par\endgroup
The hidden state starts at $r_0=0$ for each sequence and $s_t$ is repeated at every step, so both motion and hand predictions depend on the preceding commands of both groups. The energy combines mean squared prediction errors for motion and hand commands:
\begin{align}
 E_m&=\frac{\sum_{k=1}^{h-1}\norm{m_k-\tilde m_k}_2^2}{(h-1)d_m},\nonumber\\
 E_u&=\frac{\sum_{k=1}^{h-1}\norm{u_k-\tilde u_k}_2^2}{(h-1)d_u},\nonumber\\
 E_\phi(B;s_t)&=w_m E_m+w_u E_u,\quad w_m+w_u=1.
 \label{eq:energy}
\end{align}
We use $w_m=d_m/d_a$ and $w_u=d_u/d_a$ so each normalized coordinate contributes equally.

\subsection{Action Consistency Regularization}
\label{sec:inverse}
We train the endpoint predictor with endpoint feature matching and action consistency regularization. Dense visual supervision and pooled alignment are
\begin{align}
 \mathcal L_{\mathrm{wm}}&=\frac{\norm{\widehat{\mathbf f}_{t+h}-\sg(\mathbf f_{t+h})}_1}{|\mathbf f_{t+h}|},\nonumber\\
 \Delta z&=\pool(\mathbf f_{t+h})-\pool(\mathbf f_t),\nonumber\\
 \Delta\hat z&=\pool(\widehat{\mathbf f}_{t+h})-\sg(\pool(\mathbf f_t)),\nonumber\\
 \mathcal L_\Delta&=\mse(\Delta\hat z,\sg(\Delta z)).
 \label{eq:visual_losses}
\end{align}
Here $\sg$ stops gradients. The shared frozen current feature cancels in $\mathcal L_\Delta$, yielding pooled endpoint alignment.

Action consistency regularization uses an inverse action decoder $D_\psi$ to reconstruct the first demonstrated command from observed and predicted feature changes:
\begin{align}
 \mathcal L_{\mathrm{acr}}={}&\mse(D_\psi(\sg(\Delta z),s_t),a_{t,0})\nonumber\\
 &+\mse(D_\psi(\Delta\hat z,s_t),a_{t,0}).
 \label{eq:inverse_loss}
\end{align}
The observed branch trains the decoder on demonstrated transitions. The predicted branch passes command-reconstruction gradients to the predictor and its conditioning modules, encouraging predictions to retain action-relevant information.

The prior learns from demonstrations $\mathcal D$ through $\mathcal L_{\mathrm{prior}}=\E_{\mathcal D}E_\phi(B;s_t)$. Together with the action generator's flow-matching loss, the total objective is
\begin{align}
 \mathcal L={}&\mathcal L_{\mathrm{act}}+\lambda_{\mathrm{wm}}\mathcal L_{\mathrm{wm}}
 +\lambda_\Delta\mathcal L_\Delta\nonumber\\
 &+\lambda_{\mathrm{acr}}\mathcal L_{\mathrm{acr}}
 +\lambda_{\mathrm{prior}}\mathcal L_{\mathrm{prior}}.
 \label{eq:total}
\end{align}
All trainable components are optimized jointly with this objective.

\section{Experiments}\label{sec:experiments}
\subsection{Experimental Setup}
\label{sec:experimental-setup}

\begin{table}[H]
\centering\footnotesize
\caption{Platforms and implementation settings.}
\label{tab:setup}
\begin{tabularx}{\columnwidth}{@{}lX@{}}
\toprule
\textbf{Setting} & Value \\
\midrule
Real robot & Unitree G1 \\
Egocentric camera & ZED2 \\
Training hardware & $\TrainingGPUCount\times$ NVIDIA A100 \\
Evaluation (sim./real) & $1\times$ NVIDIA RTX 4090D (48 GB) \\
Onboard deployment test & Jetson Orin NX (16 GB) \\
Proposal length & Sim.: $H=30$; real: $H=40$ \\
Evaluation horizon & $h=20$ \\
Executed prefix length & $K=10$; commands at 30 Hz \\
Command chunks & $N=8$ \\
Joint temporal prior & GRU, 2 layers, hidden width 512 \\
Loss weights & $\lambda_{\mathrm{wm}}=0.1$, $\lambda_\Delta=0.05$,\newline $\lambda_{\mathrm{acr}}=0.02$, $\lambda_{\mathrm{prior}}=0.01$ \\
Score weight & $\beta=0.1$ \\
Subgoals per task & $J=5$ \\
Subgoal tracker & $\epsilon_g=0.15$ \\
\bottomrule
\end{tabularx}
\end{table}

\begin{figure*}[!b]
\centering
\includegraphics[width=\textwidth]{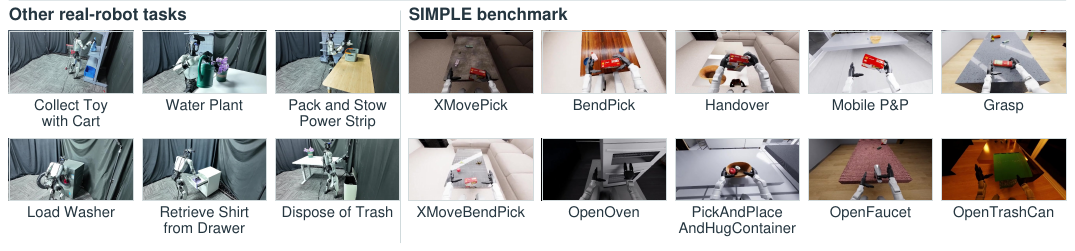}
\caption{Representative task scenes: the other six real-robot tasks and the ten SIMPLE simulation tasks. Place Basket on Table is shown in Fig.~\ref{fig:teaser}.}
\label{fig:task-gallery}
\end{figure*}

\emph{Tasks and demonstrations:} We evaluate ten tasks from SIMPLE~\cite{simple}, a simulation benchmark for humanoid loco-manipulation (Fig.~\ref{fig:task-gallery}), using its official demonstrations collected through motion planning and VR teleoperation. Its standard 36-dimensional interface contains 14 hand joint targets and 22 channels for arm, torso, and locomotion control. We run ten trials per task at each of three randomization levels: L0 varies distractors and backgrounds; L1 adds materials and lighting; L2 adds initial object and robot poses. Real-robot evaluation covers seven G1 tasks (Figs.~\ref{fig:teaser} and~\ref{fig:task-gallery}) under standard conditions and unseen object configurations, with ten trials per task and condition. The real-robot interface combines a 64-dimensional SONIC motion latent with 14 hand joint targets. Demonstrations are collected using SONIC: 400 per task, except for 200 in Place Basket on Table. Unseen configurations change object identity, placement, orientation, or appearance to states absent from the corresponding task demonstrations, while preserving the task objective.

\emph{Visual subgoals:} For offline subgoal extraction, we mean-pool all spatiotemporal tokens from each repeated-image clip using the frozen visual encoder, accumulate feature changes between consecutive frames, and sample $J$ subgoal images at evenly spaced values of this cumulative change. The images remain in temporal order, and standard and unseen-configuration trials use the same sequence for each task; the baselines do not use subgoals.

\emph{Implementation:} Table~\ref{tab:setup} summarizes the platforms and implementation settings. Each method trained in our experiments uses a separate policy per task. Simulation policies use \SimTrainingSteps{} training steps and a per-GPU batch size of \SimBatchPerGPU{}; real-robot policies use \RealTrainingSteps{} steps and a per-GPU batch size of \RealBatchPerGPU{}.

\emph{Baselines:} For the simulation comparison, we train and evaluate DiT4DiT~\cite{dit4dit}, Fast-WAM~\cite{fastwam}, and VLA-JEPA~\cite{vlajepa}; task-success results for the other baselines are taken from SIMPLE~\cite{simple}. For the real-robot comparison, we train and evaluate GR00T N1.7~\cite{gr00tn1}, $\Psi_0$~\cite{psi0}, and DiT4DiT.

\begin{figure*}[tp]
\centering
\includegraphics[width=0.92\textwidth]{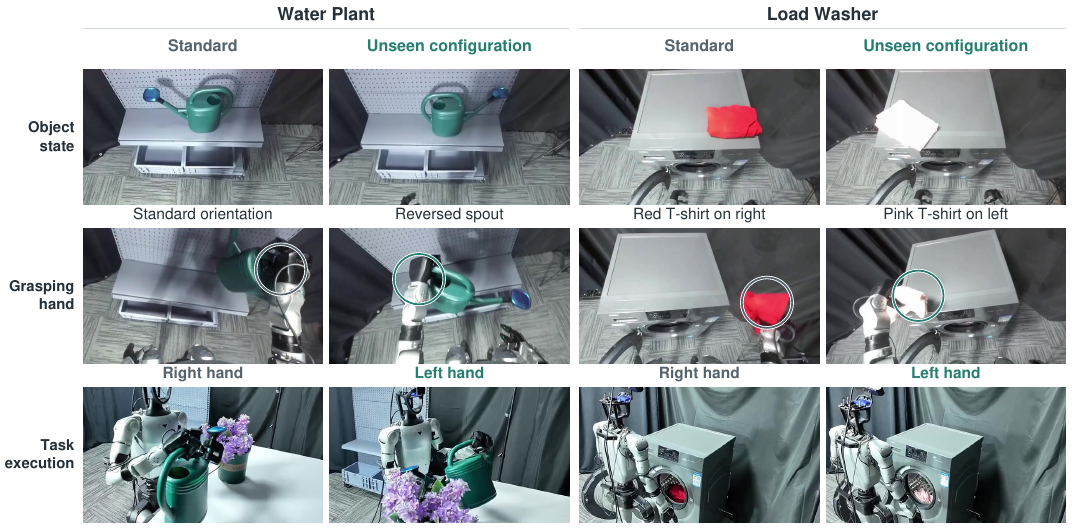}
\caption{Grasping behavior of our method under standard and unseen object configurations. Rings mark the grasping hand.}
\label{fig:hand-adaptation}
\end{figure*}

\subsection{Task Performance}
Tables~\ref{tab:simple}--\ref{tab:real-standard} report closed-loop task success.
\begin{table*}[!b]
\centering\footnotesize
\renewcommand{\alignedname}{\oursname}
\caption{Six main SIMPLE tasks: L0/L1/L2 successes per ten trials. DiT4DiT, Fast-WAM, VLA-JEPA, and Ours are evaluated by us; other rows are from SIMPLE~\cite{simple}. Task names follow SIMPLE's results table: Mobile P\&P is LocomotionPickBetweenTables, and BendPick and Grasp (TabletopGrasp) use motion-planning demonstrations. Bold/underline mark best/second-best.}
\label{tab:simple}
\setlength{\tabcolsep}{3pt}
\begin{tabular}{lccccccc}
\toprule
 & \multicolumn{6}{c}{\textbf{Main SIMPLE Tasks (L0/L1/L2)}} & \\
\cmidrule(lr){2-7}
\textbf{Method} & XMovePick & BendPick & Handover & Mobile P\&P & Grasp & XMoveBendPick & \shortstack{\textbf{Avg.}\\(\%)} \\
\midrule
\SimpleBenchmarkRows
\midrule
\EndpointSimpleRows
\bottomrule
\end{tabular}
\end{table*}

\begin{table*}[!tbp]
\centering\footnotesize
\renewcommand{\alignedname}{\oursname}
\caption{Four additional SIMPLE tasks. Results for $\Psi_0$ are from SIMPLE~\cite{simple}; we evaluate the other methods. Reporting follows Table~\ref{tab:simple}. Overall is the average success rate across all ten tasks.}
\label{tab:simple-additional}
\setlength{\tabcolsep}{3pt}
\begin{tabular}{lcccccc}
\toprule
 & \multicolumn{4}{c}{\textbf{Additional SIMPLE Tasks (L0/L1/L2)}} & & \\
\cmidrule(lr){2-5}
\textbf{Method} & OpenOven & \shortstack{PickAndPlace\\AndHugContainer} & OpenFaucet & OpenTrashCan & \shortstack{\textbf{Avg. (\%)}\\(4 tasks)} & \shortstack{\textbf{Overall (\%)}\\(10 tasks)} \\
\midrule
\SimpleAdditionalRows
\midrule
\EndpointAdditionalRows
\bottomrule
\end{tabular}
\end{table*}

\begin{table*}[!t]
\centering\footnotesize
\setlength{\tabcolsep}{4pt}
\caption{Real-robot success under standard and unseen object configurations. Cells show successes out of ten trials; Avg. is the mean success rate across tasks (\%). Bold/underline mark best/second-best.}
\label{tab:real-standard}
\label{tab:real-generalization}
\begin{tabular}{lcccccccc}
\toprule
\textbf{Method} & \shortstack{Dispose of\\Trash} & \shortstack{Collect Toy\\with Cart} & \shortstack{Water\\Plant} & \shortstack{Pack and Stow\\Power Strip} & \shortstack{Load\\Washer} & \shortstack{Retrieve Shirt\\from Drawer} & \shortstack{Place Basket\\on Table} & \textbf{Avg.} \\
\midrule
\multicolumn{9}{c}{\textbf{(a) Standard conditions}} \\
\midrule
\ConfirmedRealStandardRows
\midrule
\multicolumn{9}{c}{\textbf{(b) Unseen object configurations}} \\
\midrule
\ConfirmedRealGeneralizationRows
\bottomrule
\end{tabular}
\end{table*}

\emph{Simulation:} Our method achieves \AlignedSimMacro\% average success across ten SIMPLE tasks, compared with \SimpleDitMacro\% for DiT4DiT, \SimplePsiMacro\% for $\Psi_0$, and \SimpleVLAMacro\% for VLA-JEPA. Success averages \AlignedSimMainMacro\% on the six main tasks and \AlignedSimAdditionalMacro\% on the four additional tasks. OpenFaucet is the most difficult task, with \EndpointFaucetSuccesses/\EndpointFaucetTrials{} successes.

\emph{Real robot:} Our method achieves \AlignedRealStandardMacro\% success under standard conditions and \AlignedRealGeneralMacro\% under unseen object configurations, compared with \LayoutRealGRStandardMacro\% and \LayoutRealGRGeneralMacro\% for GR00T N1.7, the strongest baseline. Under standard conditions, our method succeeds in all ten trials on four tasks. The gain of \LayoutRealStandardGain{} percentage points comes mainly from Load Washer (10/10 versus 2/10). GR00T N1.7 failures on this task include dropping the T-shirt during transport, releasing it before it is inside the drum, and not opening the washer door. The first two resemble the Place Basket on Table failure in Fig.~\ref{fig:teaser}, where the basket falls during placement: the grasp is not maintained until the object reaches its target. Dispose of Trash is the only task where our method has a lower success rate than GR00T N1.7; its failures include the partial lid opening in Fig.~\ref{fig:failures}. Under unseen configurations, baseline success ranges from 0/10 to 6/10 per task. Our method has the highest or tied-highest success on all seven tasks and reaches at least 7/10 on four of them.

\subsection{Qualitative Results and Generalization}
\label{sec:generalization}
In Water Plant and Load Washer (Fig.~\ref{fig:hand-adaptation}), the robot uses its right hand in standard trials and its left hand in the unseen object configurations with a reversed watering-can spout or a T-shirt placed on its left, completing both tasks. The left-hand T-shirt grasp does not appear in the demonstrations.

\subsection{Data Efficiency}
\label{sec:data-efficiency}
\begin{figure}[!htbp]
\centering
\includegraphics[width=\columnwidth]{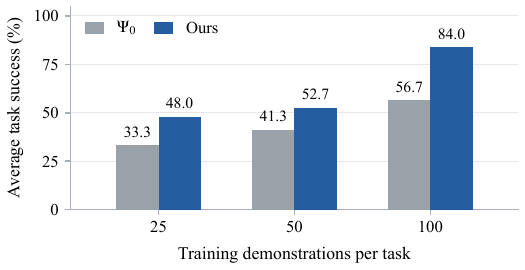}
\caption{Average success on five SIMPLE tasks versus the number of training demonstrations per task.}
\label{fig:data-efficiency}
\end{figure}

We compare training with 25, 50, and 100 demonstrations per task on five SIMPLE tasks: OpenOven, PickAndPlaceAndHugContainer, OpenFaucet, OpenTrashCan, and Mobile P\&P.

At each of the three demonstration budgets, our method has higher average success than $\Psi_0$ on this five-task subset (Fig.~\ref{fig:data-efficiency}). With 100 demonstrations per task, it reaches \DataEfficiencyOursHundred\%, compared with \DataEfficiencyPSIHundred\% for $\Psi_0$, a gain of \DataEfficiencyHundredGain{} percentage points.

\subsection{Failure Analysis}
\label{sec:failures}
\begin{figure}[tbp]
\centering
\includegraphics[width=\columnwidth]{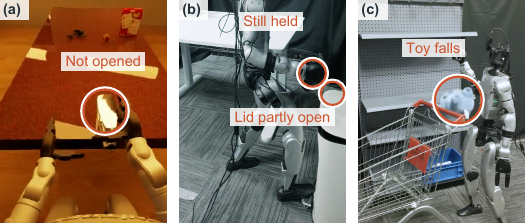}
\caption{Failure cases of our method: (a) OpenFaucet (simulation), (b) Dispose of Trash, and (c) Collect Toy with Cart.}
\label{fig:failures}
\end{figure}

OpenFaucet accounts for \EndpointFaucetFailures{} of \EndpointSimulationFailures{} simulation failures (\EndpointFaucetFailureShare\%). The robot repeatedly moves its hand near the faucet handle but fails to open it (Fig.~\ref{fig:failures}(a)). In Dispose of Trash, pressing the pedal opens the lid only partly; the robot keeps holding the object and does not release it (Fig.~\ref{fig:failures}(b)). In Collect Toy with Cart, the robot lifts the soft toy but cannot hold it securely, and the toy falls before reaching the cart (Fig.~\ref{fig:failures}(c)).

\subsection{Ablation Studies}
\label{sec:ablations}
We evaluate three design choices through closed-loop task success: prediction formulation, action consistency regularization, and temporal scoring.

\subsubsection{Prediction Formulation}
The feature-shift variant jointly encodes a short clip and predicts each visual feature block from the preceding ones. Command tokens are resampled to one group per visual block, added position-wise to the VLM prediction tokens, and interleaved with the blocks in a block-causal predictor; at inference, the current image is repeated to form the clip, and the predicted blocks are pooled for scoring. The endpoint model instead encodes current and future observations separately and predicts a single endpoint from the first $h$ commands. With joint encoding, feature-shift inputs can carry information from later frames during training that the repeated-image clip lacks at inference; separate encoding avoids this mismatch. The endpoint model reaches \AlignedSimMacro\% versus \SimpleOursMacro\% on the ten SIMPLE tasks (Table~\ref{tab:prediction}), improving both task sets.

\begin{table}[!htbp]
\centering\footnotesize
\centering\footnotesize
\caption{Prediction formulation: closed-loop success on the SIMPLE benchmark (\%).}
\label{tab:prediction}
\setlength{\tabcolsep}{4pt}
\renewcommand{\alignedname}{Endpoint (\oursname)}
\begin{tabularx}{\linewidth}{@{}p{0.37\linewidth}*{3}{>{\centering\arraybackslash}X}@{}}
\toprule
\textbf{Configuration} & \shortstack{Main\\(6 tasks)} & \shortstack{Additional\\(4 tasks)} & \shortstack{Overall\\(10 tasks)} \\
\midrule
\SimplePredictionRows
\bottomrule
\end{tabularx}

\end{table}

\emph{Component ablations:} Both ablations use the four additional SIMPLE tasks in Table~\ref{tab:simple-additional}, on which the full model has lower average success than on the six main tasks: ten complete episodes per task at each level (L0/L1/L2), totaling 120 per configuration. Within each ablation, configurations share demonstration data, training budgets, test initial conditions, and subgoals.

\subsubsection{Action Consistency Regularization}
\label{sec:ranking}
We train two models with action consistency regularization enabled or disabled, keeping the other objectives unchanged. Both use visual-only scoring at inference (Table~\ref{tab:factorial}).

\begingroup\interlinepenalty=10000
Action consistency regularization increases success from \AblationNoReconSuccesses/120 episodes (\AblationNoReconPercent\%) to \AblationVisualSuccesses/120 (\AblationVisualPercent\%), a gain of \AblationReconGain{} percentage points. Success is higher on all four tasks.
\par\endgroup

\begin{table}[!htbp]
\centering\footnotesize
\centering\footnotesize
\caption{Action consistency regularization (ACR) with visual-only scoring. Counts show successes out of ten trials at L0/L1/L2; Avg. is the success rate over all 120 episodes.}
\label{tab:factorial}
\setlength{\tabcolsep}{1.25pt}
\renewcommand{\arraystretch}{1.1}
\begin{tabularx}{\linewidth}{@{}>{\raggedright\arraybackslash}p{0.29\linewidth}*{4}{>{\centering\arraybackslash}X}>{\centering\arraybackslash}p{0.08\linewidth}@{}}
\toprule
\textbf{Configuration} & \shortstack{Open\\Oven} & \shortstack{PickAnd\\PlaceAnd\\Hug\\Container} & \shortstack{Open\\Faucet} & \shortstack{Open\\TrashCan} & \shortstack{Avg.\\(\%)} \\
\midrule
\mbox{w/o ACR} & \ClosedLoopNoReconCells \\
\mbox{w/ ACR} & \ClosedLoopVisualCells \\
\bottomrule
\end{tabularx}

\end{table}

\begingroup\interlinepenalty=10000
\subsubsection{Temporal Prior}
\label{sec:coordination}
We use the trained action generator, endpoint predictor, and joint prior from the full model, and compare alternative temporal scores at every decision (Table~\ref{tab:main}). The first- and second-order penalties average \mbox{$\|a_k-a_{k-1}\|_2^2/d_a$} and \mbox{$\|a_k-2a_{k-1}+a_{k-2}\|_2^2/d_a$}, respectively, over valid steps of the normalized command prefix. The factorized prior predicts each group from its own history and the current state; the joint prior uses the histories of both groups.
\par\endgroup

The factorized prior is trained with the other components frozen, using the joint prior's group weights and a parameter count within 1\%. All temporal scores use a fixed $\beta=0.1$; visual-only scoring sets $\beta=0$.

\begingroup\interlinepenalty=10000
The joint prior achieves the highest closed-loop success among the tested scoring methods: \AblationJointSuccesses/120 episodes (\AblationJointPercent\%). This is \AblationJointVisualGain{} percentage points above visual-only scoring and \AblationJointFactorGain{} above the factorized prior. Scoring with first- and second-order smoothness penalties achieves \AblationDiffOnePercent\% and \AblationDiffTwoPercent\% success, respectively, 1.7--4.2 percentage points above visual-only scoring, compared with 5.8 for the factorized prior. The joint prior also exceeds the factorized prior on each of the four tasks, consistent with scoring dependencies between motion and hand commands.
\par\endgroup

\begin{table}[!htbp]
\centering\footnotesize
\centering\footnotesize
\caption{Temporal-scoring ablation. Counts follow Table~\ref{tab:factorial}; the joint-prior row uses the same four-task results as Table~\ref{tab:simple-additional}.}
\label{tab:main}
\setlength{\tabcolsep}{1.25pt}
\renewcommand{\arraystretch}{1.1}
\begin{tabularx}{\linewidth}{@{}>{\raggedright\arraybackslash}p{0.29\linewidth}*{4}{>{\centering\arraybackslash}X}>{\centering\arraybackslash}p{0.08\linewidth}@{}}
\toprule
\textbf{Configuration} & \shortstack{Open\\Oven} & \shortstack{PickAnd\\PlaceAnd\\Hug\\Container} & \shortstack{Open\\Faucet} & \shortstack{Open\\TrashCan} & \shortstack{Avg.\\(\%)} \\
\midrule
\mbox{Visual only} & \ClosedLoopVisualCells \\
\mbox{1st-order smoothness} & \ClosedLoopDiffOneCells \\
\mbox{2nd-order smoothness} & \ClosedLoopDiffTwoCells \\
\mbox{Factorized prior} & \ClosedLoopFactorCells \\
\mbox{\textbf{Joint prior (Ours)}} & \ClosedLoopJointCells \\
\bottomrule
\end{tabularx}

\end{table}

\begingroup\interlinepenalty=10000
\subsection{Inference Cost and Deployment}
Our method uses less than half the GPU memory of DiT4DiT and Fast-WAM and runs faster (Table~\ref{tab:simple-efficiency}); GR00T N1.6 remains faster at similar memory use. Our task-success evaluations use the RTX 4090D listed in Table~\ref{tab:setup}. In an additional onboard deployment test, our method runs on the robot's Jetson Orin NX (16 GB), averaging \OrinMemoryMean{} GB of system memory and an inference frequency of \OrinHz{} Hz across seven tasks.
\par

\endgroup
\begin{table}[!htbp]
\centering\footnotesize
\caption{Inference cost on SIMPLE. Hz denotes command chunks per second; timing for Ours includes sequence generation, visual prediction, and prior scoring.}
\label{tab:simple-efficiency}
\setlength{\tabcolsep}{3pt}
\renewcommand{\alignedname}{\oursname}
\begin{tabular}{@{}lccc@{}}
\toprule
\textbf{Method} & \shortstack{GPU memory\\(GB) $\downarrow$} & \shortstack{Trainable params.\\(B)} & \shortstack{Inference\\(Hz) $\uparrow$} \\
\midrule
\SimpleEfficiencyBaselineRows
\midrule
\EndpointEfficiencyRows
\bottomrule
\end{tabular}

\end{table}

\section{Conclusion and Future Work}\label{sec:conclusion}
Action-conditioned latent foresight scores generated motion--hand sequences by predicted visual progress and temporal consistency before execution. It achieves \AlignedSimMacro\% success in simulation and \AlignedRealStandardMacro\% and \AlignedRealGeneralMacro\% on the real robot under standard and unseen object configurations. In closed-loop ablations, action consistency regularization adds \AblationReconGain{} percentage points and the joint prior exceeds the factorized prior by \AblationJointFactorGain{} percentage points.

The failure cases include incomplete opening of articulated mechanisms and loss of grasp on a soft object. The visual score also relies on task-specific demonstration subgoal images. In the onboard deployment test, inference runs at \OrinHz{}~Hz. Future work will add contact and object-state supervision to latent prediction and extend temporal scoring across consecutive command chunks.

{\small\bibliographystyle{IEEEtran}\bibliography{main}}
\ifdefined\papersupplement
\clearpage
\onecolumn
\appendices
\setcounter{table}{0}
\renewcommand{\thetable}{S\arabic{table}}
\renewcommand{\theHtable}{supp.\arabic{table}}
\setcounter{figure}{0}
\renewcommand{\thefigure}{S\arabic{figure}}
\renewcommand{\theHfigure}{supp.\arabic{figure}}
\supplementsection{Unseen object configurations}
\begin{table}[h]
\centering\small
\caption{Unseen object configurations.}
\renewcommand{\arraystretch}{1.25}
\begin{tabularx}{\linewidth}{@{}p{0.27\linewidth}X@{}}
\toprule
Task & Configurations \\
\midrule
Dispose of Trash & Tissue packet; tea bottle. \\
Collect Toy with Cart & Two arrangements of the blue target toy and two neighboring distractors. \\
Water Plant & Watering can shifted left; shifted right; spout reversed. \\
Pack and Stow Power Strip & Power strip rotated; repositioned on the tabletop. \\
Load Washer & Purple or pink T-shirt, placed on either the right or left side of the washer top. \\
Retrieve Shirt from Drawer & Target T-shirt repositioned; gray T-shirt nearby; pink T-shirt nearby. \\
Place Basket on Table & Blue-rimmed basket replacing the red-rimmed basket. \\
\bottomrule
\end{tabularx}
\end{table}

\clearpage
\section{Visual Subgoal Extraction}
\label{app:visual-subgoals}
\begin{center}
\includegraphics[width=\textwidth,height=0.87\textheight,keepaspectratio]{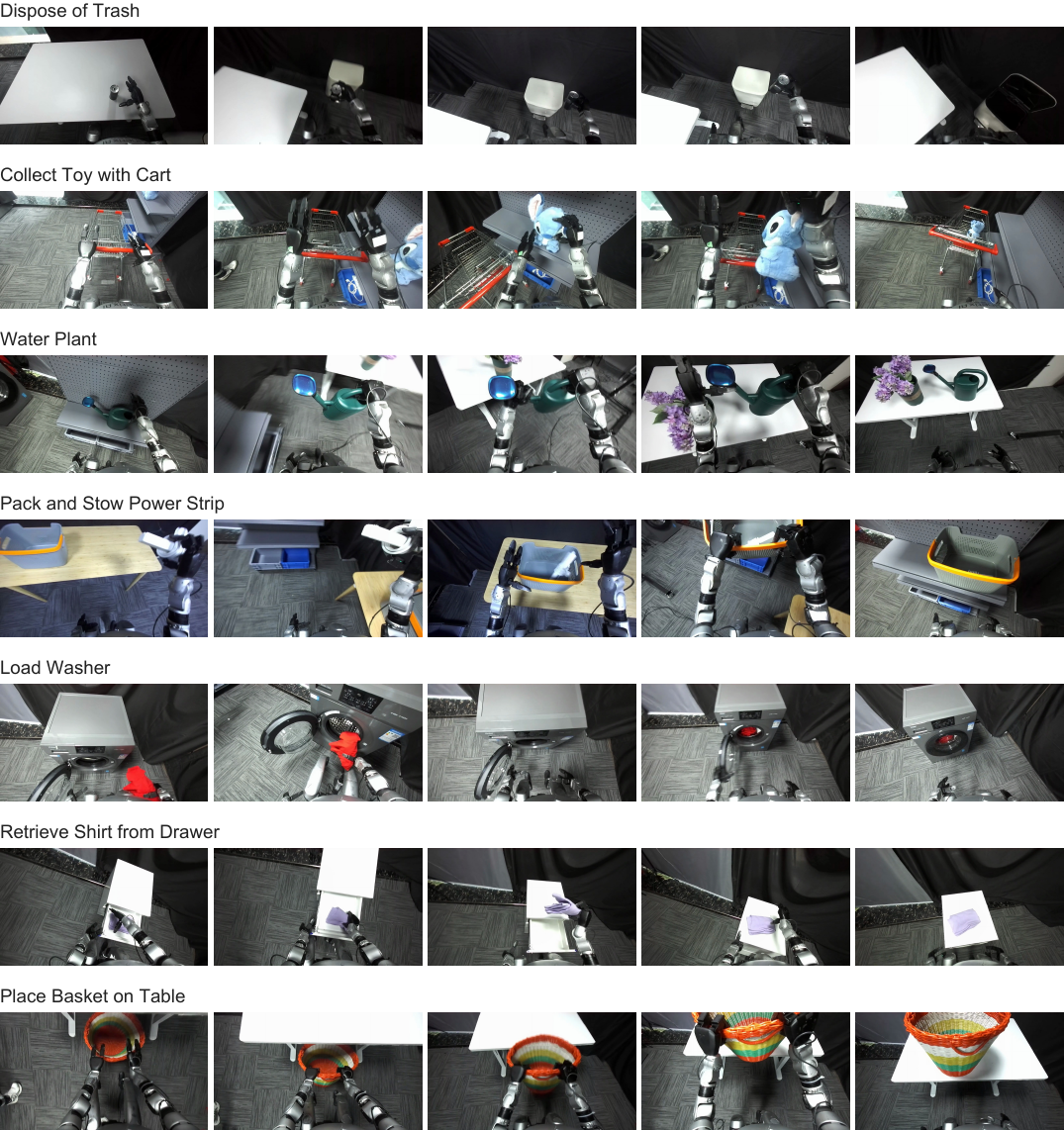}
\captionof{figure}{Visual subgoals extracted with the HIW-500-fine-tuned V-JEPA 2.1 encoder. Each row shows five temporally ordered subgoals from one task demonstration.}
\label{fig:visual-subgoals}
\end{center}

\fi
\end{document}